\documentclass[11pt,letterpaper]{article}
\usepackage{emnlp2017}
\usepackage{times}
\usepackage{latexsym}
\usepackage{amsmath}
\usepackage{amssymb}
\usepackage[pdftex]{graphicx} 

\DeclareMathOperator*{\softmax}{softmax}
\DeclareMathOperator*{\mean}{mean}

\emnlpfinalcopy

\def\emnlppaperid{***}

\title{Towards Neural Machine Translation with Latent Tree Attention}

\author{James Bradbury \and Richard Socher \\
  {\tt james.bradbury@salesforce.com} \\
  {\tt rsocher@salesforce.com}}

\date{}

\begin{document}
\maketitle
\begin{abstract}
Building models that take advantage of the hierarchical structure of language without \emph{a priori} annotation is a longstanding goal in natural language processing. 
We introduce such a model for the task of machine translation, pairing a recurrent neural network grammar encoder with a novel attentional RNNG decoder and applying policy gradient reinforcement learning to induce unsupervised tree structures on both the source and target. When trained on character-level datasets with no explicit segmentation or parse annotation, the model learns a plausible segmentation and shallow parse, obtaining performance close to an attentional baseline.
%
%
\end{abstract}

\section{Introduction}


Many efforts to exploit linguistic hierarchy in NLP tasks
make use of the output of a self-contained parser system trained from a human-annotated treebank \citep{Huang2006}. An alternative approach aims to jointly learn the task at hand and relevant aspects of linguistic hierarchy, inducing from an unannotated training dataset parse trees that may or may not correspond to treebank annotation practices \citep{Wu1997, Chiang2005}.
%
%
%
%
%
%
%

Most deep learning models for NLP that aim to make use of linguistic hierarchy integrate an external parser, either to prescribe the recursive structure of the neural network \citep{Pollack1990, Goller1996, Socher2013} or to provide a supervision signal or training data for a network that predicts its own structure \citep{Socher2010, Bowman2016, Dyer2016b}. But some recently described neural network models take the second approach and treat hierarchical structure as a latent variable, applying inference over graph-based conditional random fields \citep{Kim2017}, the straight-through estimator \citep{Chung2017}, or policy gradient reinforcement learning \citep{Yogatama2017} to work around the inapplicability of gradient-based learning to problems with discrete latent states.
%
%

For the task of machine translation,
syntactically-informed models have shown promise both inside and outside the deep learning context, with hierarchical phrase-based models frequently outperforming traditional ones \citep{Chiang2005} and neural MT models augmented with morphosyntactic input features \citep{Sennrich2016, Nadejde2017}, a tree-structured encoder \citep{Eriguchi2016, Hashimoto2017}, and a jointly trained parser \citep{Eriguchi2017} each outperforming purely-sequential baselines.

Drawing on many of these precedents, we introduce an attentional neural machine translation model whose encoder and decoder components are both tree-structured neural networks that predict their own constituency structure as they consume or emit text. The encoder and decoder networks are variants of the RNNG model introduced by \citet{Dyer2016b}, allowing tree structures of unconstrained arity, while text is ingested at the character level, allowing the model to discover and make use of structure within words.

The parsing decisions of the encoder and decoder RNNGs are parameterized by a stochastic policy trained using a weighted sum of two objectives: a language model loss term that rewards predicting the next character with high likelihood, and a tree attention term that rewards one-to-one attentional correspondence between constituents in the encoder and decoder.

We evaluate this model on the German-English language pair of the \texttt{flickr30k} dataset, where it obtains similar performance to a strong character-level baseline. Analysis of the latent trees produced by the encoder and decoder shows that the model learns a reasonable segmentation and shallow parse, and most phrase-level constituents constructed while ingesting the German input sentence correspond meaningfully to constituents built while generating the English output.

\section{Model}

\subsection{Encoder/Decoder Architecture}

The model consists of a coupled encoder and decoder, where the encoder is a modified stack-only recurrent neural network grammar \citep{Kuncoro2017} and the decoder is a stack-only RNNG augmented with constituent-level attention. An RNNG is a top-down transition-based model that jointly builds a sentence representation and parse tree, representing the parser state with a StackLSTM and using a bidirectional LSTM as a constituent composition function.
\begin{figure}[t!]
\centering
\includegraphics[width=.99\linewidth]{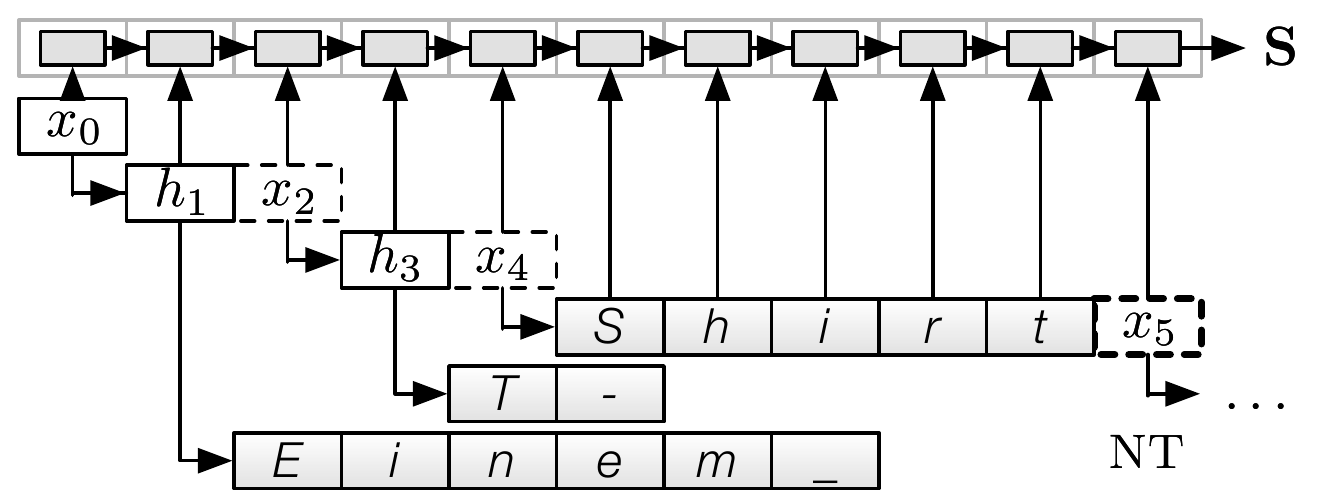}
\includegraphics[width=.99\linewidth]{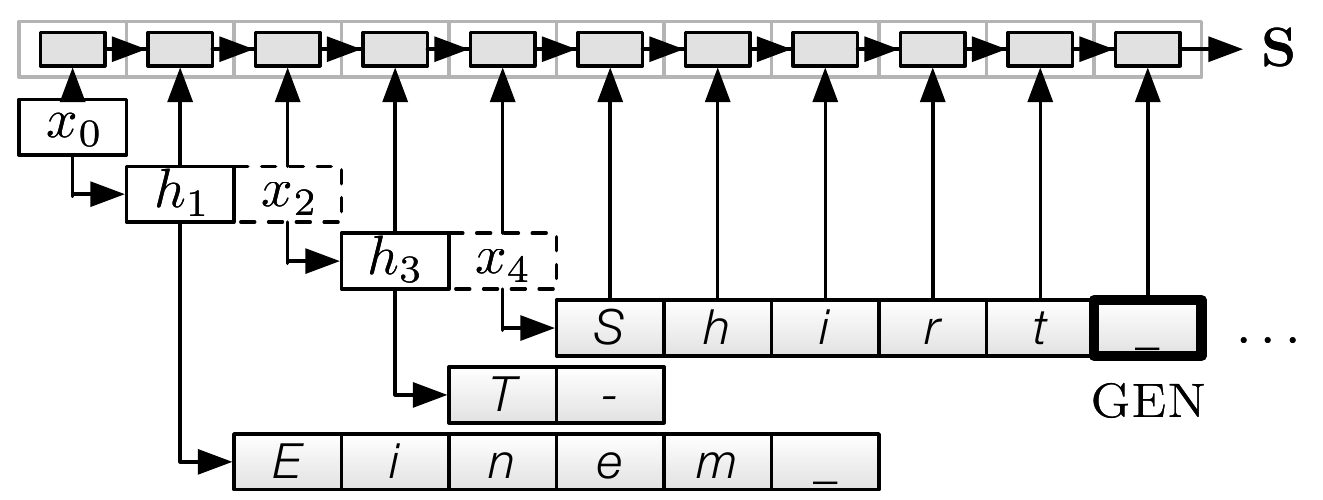}
\includegraphics[width=.99\linewidth]{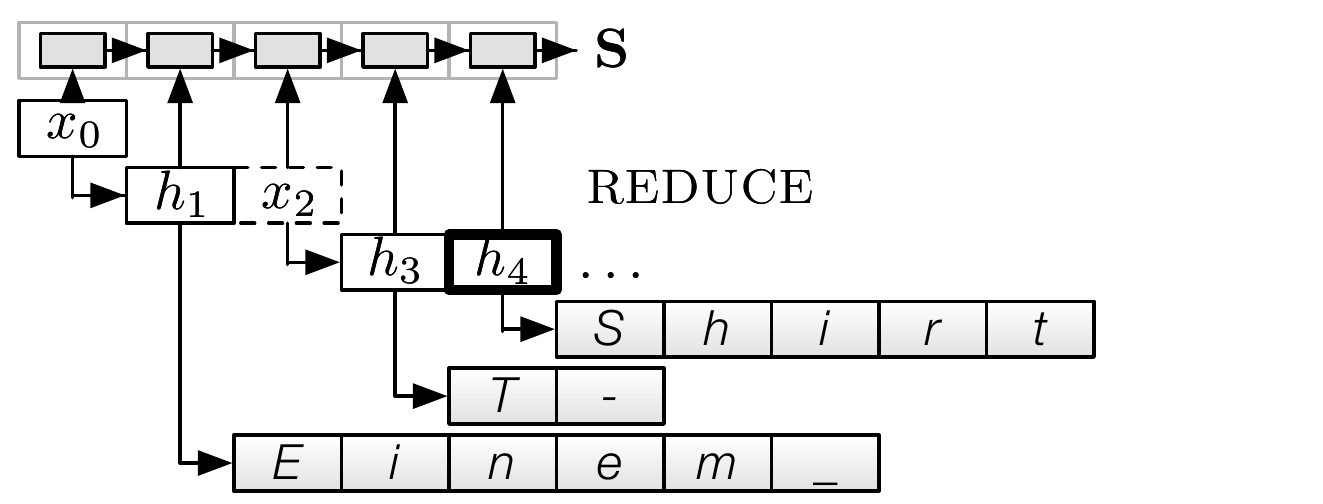}
\caption{
At a given timestep during either encoding or decoding
there are three possible transitions (although one or more may be forbidden): begin a new nonterminal constituent (\textsc{nt}), predict and ingest a terminal (\textsc{gen}), or end the current nonterminal (\textsc{reduce}). If the chosen transition is \textsc{nt}, the RNNG adds a new-nonterminal token $\mathbf{x}_{i+1}$ to the active constituent and begins a new nonterminal constituent (\textbf{1a}). If the transition is \textsc{gen}, the RNNG predicts the next token (Section 2.3) and adds the ground-truth next token $\mathbf{e}$ from the context buffer at the cursor location (\textbf{1b}). If the transition is \textsc{reduce}, the contents of the active nonterminal are passed to the composition function, the new-nonterminal token $\mathbf{x}_i$ is replaced with the result of the composition $\mathbf{h}_i$, and the StackLSTM rolls back to the previously active constituent (\textbf{1c}). In all three cases, the StackLSTM then advances one step with the newly added token as input ($\mathbf{x}_{i+1}$, $\mathbf{e}$, or $\mathbf{h}_i$).
}
\label{fig:RNNG}
\end{figure}
Our implementation is detailed in Figure 1, and differs from \citet{Dyer2016b} in that it lacks separate new-nonterminal tokens for different phrase types, and thus does not include the phrase type as an input to the composition function. Instead, the values of $\mathbf{x}_i$ for the encoder are fixed to a constant $\mathbf{x}^{\rm enc}$ while the values of $\mathbf{x}_j$ for the decoder are determined through an attention procedure (Section 2.2).

As originally described, the RNNG predicts parser transitions using a one-layer $\tanh$ perceptron with three concatenated inputs: the last state of a unidirectional LSTM over the stack contents ($\mathbf{s}$), the last state of a unidirectional LSTM over the reversed buffer of unparsed tokens ($\mathbf{b}$), and the result of an LSTM over the past transitions ($\mathbf{a}$). All three of these states can be computed with at most one LSTM step per parser transition using the StackLSTM algorithm \citep{Dyer2016a}. But such a baseline RNNG is actually outperformed by one which conditions the parser transitions only on the stack representation \citep{Kuncoro2017}. Restricting our model to this stack-only case allows both the encoder and decoder to be supervised using a language model loss, while allowing the model access to $\mathbf{b}$ would give it a trivial way to predict the next character and obtain zero loss.

\subsection{Attention}

With the exception of the attention mechanism, the encoder and decoder are identical. While the encoder uses a single token to represent a new nonterminal, the decoder represents a new nonterminal on the stack as a sum weighted by \emph{structural attention} of the \emph{phrase representations} of all nonterminal tree nodes produced by the encoder. In particular, we use the normalized dot products between the decoder stack representation $\mathbf{s}_j^{\rm dec}$ and the stack representation at each encoder node $\mathbf{s}^i_{\rm enc}$ (that is, the hidden state of the StackLSTM up to and including $\mathbf{x}_j^{\rm enc}$ but not $\mathbf{h}_j^{\rm enc}$) as coefficients in a weighted sum of the phrase embeddings $\mathbf{h}^i_{\rm enc}$ corresponding to the encoder nodes:
\begin{align}
\begin{split}\label{attn}
\alpha_{ij}&=\softmax_{\text{all }i}(\mathbf{s}_i^{\rm enc}\cdot\mathbf{s}_j^{\rm dec})\\
\mathbf{x}_j^{\rm dec}&=\sum_i\alpha_{ij}\mathbf{h}_i^{\rm enc}.\\
\end{split}
\end{align}
Since the dot products between encoder and decoder stack representations are a measure of structural similarity between the (left context of) the current decoder state and the encoder state.
Within a particular decoder nonterminal, the model reduces to ordinary sequence-to-sequence transduction. Starting from the encoder's representation of the corresponding nonterminal or a weighted combination of such representations, the decoder will emit a translated sequence of child constituents (both nonterminal and terminal) one by one---applying attention only when emitting nonterminal children.

\subsection{Training}

We formulate our model as a stochastic computation graph \citep{Schulman2015}, leading to a training paradigm that combines backpropagation (which provides the exact gradient through deterministic nodes) and vanilla policy gradient (which provides a Monte Carlo estimator for the gradient through stochastic nodes).

There are several kinds of training signals in our model. First, when the encoder or decoder chooses the \textsc{gen} action it passes the current stack state $\mathbf{s}$ through a one-layer softmax perceptron, giving the probability that the next token is each of the characters in the vocabulary.
The language model loss $\mathcal{L}_k$ for each generated token is the negative log probability assigned to the ground-truth next token. The other differentiable training signal is the coverage loss $\mathcal{L}_c$, which is a measure of how much the attention weights diverge from the ideal of a one-to-one mapping. This penalty is computed as a sum of three MSE terms:
\begin{align}
\begin{split}\label{coverage}
\mathcal{L}_c&=\mean_{\text{all }i}(1 - \sum_{\text{all }j}\alpha_{ij})^2\\
&+\mean_{\text{all }i}(1 - \max_{\text{all }j}\alpha_{ij})^2\\
&+\mean_{\text{all }j}(1 - \max_{\text{all }i}\alpha_{ij})^2\\
\end{split}
\end{align}
Backpropagation using the differentiable losses affects only the weights of the output softmax perceptron. The overall loss function for these weights is a weighted sum of all $\mathcal{L}_k$ terms and $\mathcal{L}_c$:
\begin{align}
\begin{split}\label{loss}
\mathcal{L}&=100\mathcal{L}_c+10\sum_{\text{all }k}\mathcal{L}_k
\end{split}
\end{align}
There are additionally nondifferentiable rewards $r$ that bias the model towards or away from certain kinds of tree structures. Here, negative numbers correspond to penalties. We assign a tree reward of $-1$ when the model predicts a \textsc{reduce} with only one child constituent (\textsc{reduce} with zero child constituents is forbidden) or predicts two \textsc{reduce} or \textsc{nt} transitions in a row. This biases the model against unary branching and reduces its likelihood of producing an exclusively left- or right-branching tree structure. In addition, for all constituents except the root, we assign a tree reward based on the size and type of its children. If $n$ and $t$ are the number of nonterminal and terminal children, this reward is $4t$ if all children are terminal and $9\sqrt{n}$ otherwise. A reward structure like this
biases the model against freely mixing terminals and nonterminals within the same constituent and provides incentive to build substantial tree structures early on in training so the model doesn't get stuck in trivial local minima.

Within both the encoder and decoder, each stochastic action node has a corresponding tree reward $r_k$ if the action was \textsc{reduce} (otherwise zero) and a corresponding language model loss $\mathcal{L}_k$ if the action was \textsc{gen} (otherwise zero). We subtract an exponential moving average baseline from each tree reward and additional exponential moving average baselines---computed independently for each character $z$ in the vocabulary, because we want to reduce the effect of character frequency---from the language model losses. If $\textsc{gen}(k)$ is the number of \textsc{gen} transitions among actions one through $k$, and $\gamma$ is a decay constant, the final reward $\mathcal{R}_k^m$ for action $k$ with $m\in\{\text{enc}, \text{dec}\}$ is:
\begin{align}
\begin{split}\label{encoder}
\hat{r}_k&=r_k-r_{\rm baseline}\\
\hat{\mathcal{L}}_k&=\mathcal{L}_k-\mathcal{L}_{\rm baseline}(z_k)\\
\hat{\mathcal{R}}_k&=\sum_{\kappa=k}^{K_m}\gamma^{\textsc{gen}(\kappa)-\textsc{gen}(k)}(\hat{r}_\kappa-\hat{\mathcal{L}}_\kappa^m)\\
\mathcal{R}_k^{\rm m}&=\hat{\mathcal{R}}_k-\mathcal{L}_c-(m=\text{enc})\sum_{\kappa=1}^{K_{\rm dec}}\mathcal{L}_k^{\rm dec}.
\end{split}
\end{align}
These rewards define the gradient that each stochastic node (with normalized action probabilities $p_k^a$ and chosen action $a_k$) produces during backpropagation according to the standard multinomial score function estimator (REINFORCE):
\begin{align}
\begin{split}\label{reinforce}
\nabla_{\theta} p_k^a &=
\mean_{\rm a_k=a}\mathcal{R}_k \nabla_{\theta} \log p_k^{a_k}=
\mean_{\rm a_k=a}\frac{-\mathcal{R}_k}{p_k^{a_k}}
\end{split}
\end{align}
\section{Results}
We evaluated our model on the German-English language pair of the \texttt{flickr30k} data, the textual component of the WMT Multimodal Translation shared task \citep{Specia2016}. An attentional sequence-to-sequence model with two layers and 384 hidden units from the OpenNMT project \citep{Klein2017} was run at the character level as a baseline, obtaining 32.0 test BLEU with greedy inference. Our model with the same hidden size and greedy inference achieves test BLEU of 28.5 after removing repeated bigrams.

We implemented the model in PyTorch, benefiting from its strong support for dynamic and stochastic computation graphs, and trained with batch size 10 and the Adam optimizer \citep{Kingma2015} with early stopping after 12 epochs. Character embeddings and the encoder's $\mathbf{x}^{\rm enc}$ embedding were initialized to random 384-dimensional vectors. The value of $\gamma$ and the decay constant for the baselines' exponential moving average were both set to 0.95.
\begin{figure}[t!]
\centering
\includegraphics[width=.99\linewidth]{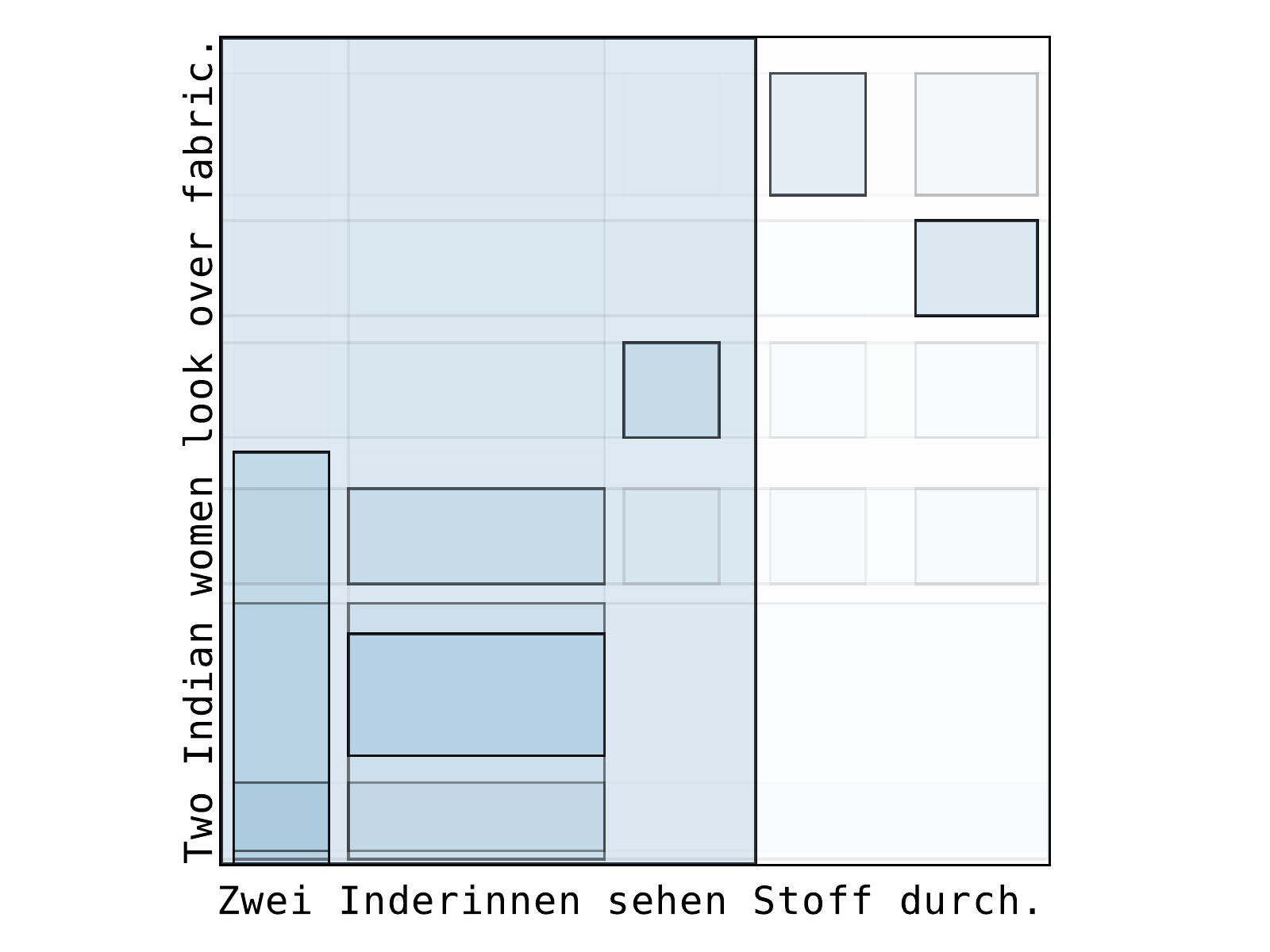}
\includegraphics[width=.99\linewidth]{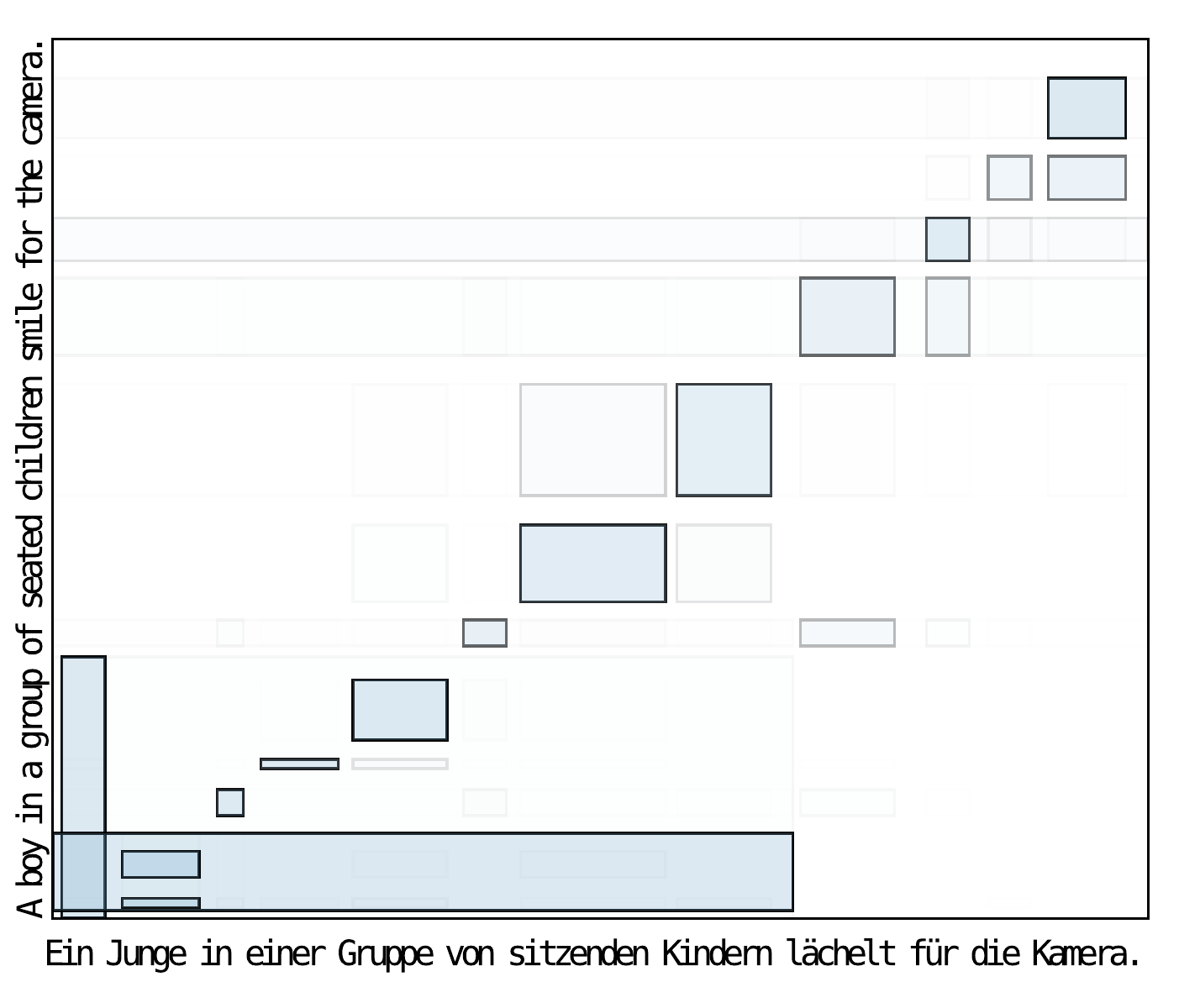}
\caption{
Attention visualizations for two sentences from the development set. Attention between two constituents is represented by a shaded rectangle whose projections on the $x$ and $y$ axes cover the encoder and decoder constituents respectively.
}
\label{fig:attention}
\end{figure}
A random selection of translations is included in the supplemental material, while two attention plots are shown in Figure 2. Figure 2b demonstrates a common pathology of the model, where a phrasal encoder constituent would be attended to during decoding of the head word of the corresponding decoder constituent, while the head word of the encoder constituent would be attended to during decoding of the decoder constituent corresponding to the whole phrase. Another common pathology is repeated sentence fragments in the translation, which are likely generated because the model cannot condition future attention directly on past attention weights (the ``input feeding'' approach introduced by \citet{Luong2015}).

Translation quality also suffers because of our use of a stack-only RNNG, which we chose because an RNNG with both stack and buffer inputs is incompatible with a language model loss. During encoding, the model must decide at the very beginning of the sentence how deeply to embed the first character. But with a stack-only RNNG, it must make this decision randomly, since it isn't able to use the buffer representation---which contains the entire sentence.
\section{Conclusion}
We introduce a new approach to leveraging unsupervised tree structures in NLP tasks like machine translation. Our experiments demonstrate that a small-scale MT dataset contains sufficient training signal to infer latent linguistic structure, and we are excited to learn what models like the one presented here can discover in full-size translation corpora. One particularly promising avenue of research is to leverage the inherently compositional phrase representations $\mathbf{h}_i^{\rm enc}$ produced by the encoder for other NLP tasks.

There are also many possible directions for improving the model itself and the training process. Value function baselines can replace exponential moving averages, pure reinforcement learning can replace teacher forcing, and beam search can be used in place of greedy inference. Solutions to the translation pathologies presented in Section 3 are likely more complex, although one possible approach would leverage variational inference using a teacher model that can see the buffer and helps train a stack-only student model.

\newpage
\bibliography{emnlp2017}

\begin{thebibliography}{}
\expandafter\ifx\csname natexlab\endcsname\relax\def\natexlab#1{#1}\fi

\bibitem[{Bowman et~al.(2016)Bowman, Gauthier, Rastogi, Gupta, Manning, and
  Potts}]{Bowman2016}
Samuel Bowman, Jon Gauthier, Abhinav Rastogi, Raghav Gupta, Christopher
  Manning, and Christopher Potts. 2016.
\newblock A fast unified model for parsing and sentence understanding.
\newblock In {\em ACL\/}.

\bibitem[{Chiang(2005)}]{Chiang2005}
David Chiang. 2005.
\newblock A hierarchical phrase-based model for statistical machine
  translation.
\newblock In {\em ACL\/}.

\bibitem[{Chung et~al.(2017)Chung, Ahn, and Bengio}]{Chung2017}
Junyoung Chung, Sungjin Ahn, and Yoshua Bengio. 2017.
\newblock Hierarchical multiscale recurrent neural networks.
\newblock In {\em ICLR\/}.

\bibitem[{Dyer et~al.(2016{\natexlab{a}})Dyer, Ballesteros, Ling, Matthews, and
  Smith}]{Dyer2016a}
Chris Dyer, Miguel Ballesteros, Wang Ling, Austin Matthews, and Noah~A Smith.
  2016{\natexlab{a}}.
\newblock Transition-based dependency parsing with stack long short-term
  memory.
\newblock In {\em EMNLP\/}.

\bibitem[{Dyer et~al.(2016{\natexlab{b}})Dyer, Kuncoro, Ballesteros, and
  Smith}]{Dyer2016b}
Chris Dyer, Adhiguna Kuncoro, Miguel Ballesteros, and Noah Smith.
  2016{\natexlab{b}}.
\newblock Recurrent neural network grammars.
\newblock In {\em NAACL\/}.

\bibitem[{Eriguchi et~al.(2016)Eriguchi, Hashimoto, and
  Tsuruoka}]{Eriguchi2016}
Akiko Eriguchi, Kazuma Hashimoto, and Yoshimasa Tsuruoka. 2016.
\newblock Tree-to-sequence attentional neural machine translation.
\newblock In {\em ACL\/}.

\bibitem[{Eriguchi et~al.(2017)Eriguchi, Tsuruoka, and Cho}]{Eriguchi2017}
Akiko Eriguchi, Yoshimasa Tsuruoka, and Kyunghyun Cho. 2017.
\newblock Learning to parse and translate improves neural machine translation.
\newblock In {\em ACL\/}.

\bibitem[{Goller and K\"{u}chler(1996)}]{Goller1996}
Christoph Goller and Andreas K\"{u}chler. 1996.
\newblock Learning task-dependent distributed representations by
  backpropagation through structure.
\newblock In {\em IEEE International Conference on Neural Networks\/}. IEEE,
  volume~1, pages 347--352.

\bibitem[{Hashimoto and Tsuruoka(2017)}]{Hashimoto2017}
Kazuma Hashimoto and Yoshimasa Tsuruoka. 2017.
\newblock Neural machine translation with source-side latent graph parsing.
\newblock In {\em EMNLP\/}.

\bibitem[{Huang et~al.(2006)Huang, Knight, and Joshi}]{Huang2006}
Liang Huang, Kevin Knight, and Aravind Joshi. 2006.
\newblock A syntax-directed translator with extended domain of locality.
\newblock In {\em CHPJI-NLP\/}. Association for Computational Linguistics.

\bibitem[{Kim et~al.(2017)Kim, Denton, Hoang, and Rush}]{Kim2017}
Yoon Kim, Carl Denton, Luong Hoang, and Alexander Rush. 2017.
\newblock Structured attention networks.
\newblock In {\em ICLR\/}.

\bibitem[{Kingma and Ba(2015)}]{Kingma2015}
Diederik Kingma and Jimmy Ba. 2015.
\newblock Adam: {A} method for stochastic optimization.
\newblock In {\em ICLR\/}.

\bibitem[{{Klein} et~al.(2017){Klein}, {Kim}, {Deng}, {Senellart}, and
  {Rush}}]{Klein2017}
G.~{Klein}, Y.~{Kim}, Y.~{Deng}, J.~{Senellart}, and A.~M. {Rush}. 2017.
\newblock {OpenNMT}: {O}pen-source toolkit for neural machine translation.
\newblock {\em ArXiv preprint arXiv:1701.02810\/} .

\bibitem[{Kuncoro et~al.(2017)Kuncoro, Ballesteros, Kong, Dyer, Neubig, and
  Smith}]{Kuncoro2017}
Adhiguna Kuncoro, Miguel Ballesteros, Lingpeng Kong, Chris Dyer, Graham Neubig,
  and Noah Smith. 2017.
\newblock What do recurrent neural network grammars learn about syntax?
\newblock In {\em EACL\/}.

\bibitem[{Luong et~al.(2015)Luong, Pham, and Manning}]{Luong2015}
Minh-Thang Luong, Hieu Pham, and Christopher~D Manning. 2015.
\newblock Effective approaches to attention-based neural machine translation.
\newblock In {\em EMNLP\/}.

\bibitem[{Nadejde et~al.(2017)Nadejde, Reddy, Sennrich, Dwojak,
  Junczys-Dowmunt, Koehn, and Birch}]{Nadejde2017}
Maria Nadejde, Siva Reddy, Rico Sennrich, Tomasz Dwojak, Marcin
  Junczys-Dowmunt, Philipp Koehn, and Alexandra Birch. 2017.
\newblock Syntax-aware neural machine translation using {CCG}.
\newblock {\em arXiv preprint arXiv:1702.01147\/} .

\bibitem[{Pollack(1990)}]{Pollack1990}
Jordan~B Pollack. 1990.
\newblock Recursive distributed representations.
\newblock {\em Artificial Intelligence\/} 46(1):77--105.

\bibitem[{Schulman et~al.(2015)Schulman, Heess, Weber, and
  Abbeel}]{Schulman2015}
John Schulman, Nicolas Heess, Theophane Weber, and Pieter Abbeel. 2015.
\newblock Gradient estimation using stochastic computation graphs.
\newblock In {\em NIPS\/}.

\bibitem[{Sennrich and Haddow(2016)}]{Sennrich2016}
Rico Sennrich and Barry Haddow. 2016.
\newblock Linguistic input features improve neural machine translation.
\newblock In {\em WMT\/}.

\bibitem[{Socher et~al.(2010)Socher, Manning, and Ng}]{Socher2010}
Richard Socher, Christopher Manning, and Andrew Ng. 2010.
\newblock Learning continuous phrase representations and syntactic parsing with
  recursive neural networks.
\newblock In {\em NIPS Workshop on Deep Learning and Unsupervised Feature
  Learning\/}.

\bibitem[{Socher et~al.(2013)Socher, Perelygin, Wu, Chuang, Manning, Ng, and
  Potts}]{Socher2013}
Richard Socher, Alex Perelygin, Jean Wu, Jason Chuang, Christopher Manning,
  Andrew Ng, and Christopher Potts. 2013.
\newblock Recursive deep models for semantic compositionality over a sentiment
  treebank.
\newblock In {\em EMNLP\/}.

\bibitem[{Specia et~al.(2016)Specia, Frank, Sima’an, and
  Elliott}]{Specia2016}
Lucia Specia, Stella Frank, Khalil Sima’an, and Desmond Elliott. 2016.
\newblock A shared task on multimodal machine translation and crosslingual
  image description.
\newblock In {\em WMT\/}.

\bibitem[{Wu(1997)}]{Wu1997}
Dekai Wu. 1997.
\newblock Stochastic inversion transduction grammars and bilingual parsing of
  parallel corpora.
\newblock {\em Computational linguistics\/} 23(3):377--403.

\bibitem[{Yogatama et~al.(2017)Yogatama, Blunsom, Dyer, Grefenstette, and
  Ling}]{Yogatama2017}
Dani Yogatama, Phil Blunsom, Chris Dyer, Edward Grefenstette, and Wang Ling.
  2017.
\newblock Learning to compose words into sentences with reinforcement learning.
\newblock In {\em ICLR\/}.

\end{thebibliography}
\bibliographystyle{emnlp_natbib}

\includegraphics[width=2\linewidth]{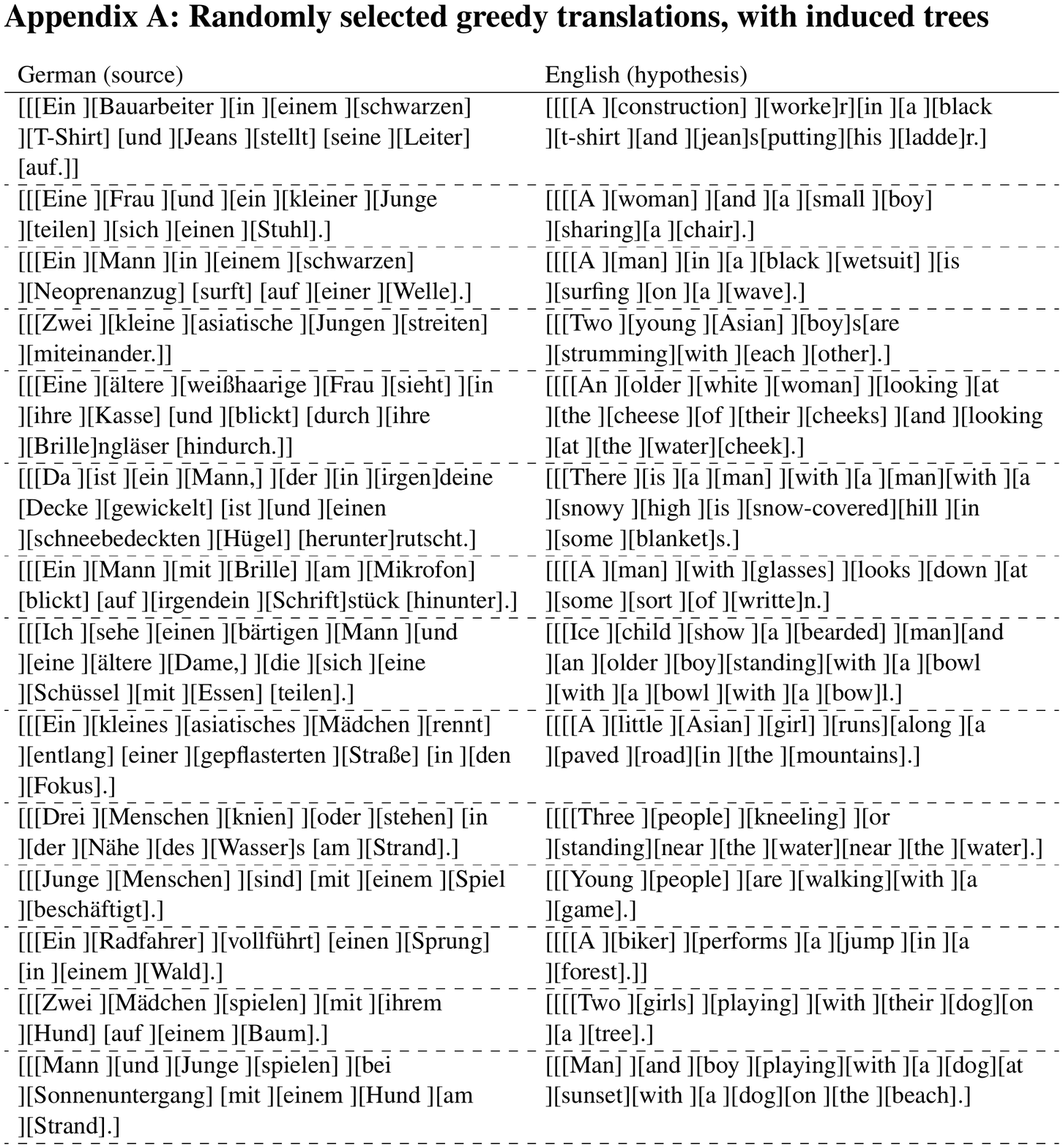}
\end{document}